\documentclass[conference]{IEEEtran}
\IEEEoverridecommandlockouts
\usepackage{cite}
\usepackage{amsmath,amssymb,amsfonts}
\usepackage{algorithm}
\usepackage{graphicx}
\usepackage{textcomp}
\def\BibTeX{{\rm B\kern-.05em{\sc i\kern-.025em b}\kern-.08em
    T\kern-.1667em\lower.7ex\hbox{E}\kern-.125emX}}

\usepackage{booktabs} 
\usepackage{makecell}
\usepackage{multirow}
\usepackage{tabularx}
\usepackage[table]{xcolor}
\usepackage{cleveref}
\usepackage{algpseudocode}

\begin{document}

\title{Adaptive Training Meets Progressive Scaling: Elevating Efficiency in Diffusion Models}

\author{Wenhao Li\textsuperscript{1}, Xiu Su\textsuperscript{2*}, Yu Han\textsuperscript{3}, Shan You\textsuperscript{4}, Tao Huang\textsuperscript{1}, Chang Xu\textsuperscript{1}\thanks{\textsuperscript{*} corresponding author}\\
\textsuperscript{1}University of Sydney, \textsuperscript{2}Central South University,
\textsuperscript{3}Nanjing Forestry University,
\textsuperscript{4}Sensetime Research}

\maketitle

\begin{abstract}
Diffusion models have demonstrated remarkable efficacy in various generative tasks with the predictive prowess of denoising model. Currently, diffusion models employ a uniform denoising model across all timesteps. However, the inherent variations in data distributions at different timesteps lead to conflicts during training, constraining the potential of diffusion models. To address this challenge, we propose a novel two-stage divide-and-conquer training strategy termed TDC Training. It groups timesteps based on task similarity and difficulty, assigning highly customized denoising models to each group, thereby enhancing the performance of diffusion models. While two-stage training avoids the need to train each model separately, the total training cost is even lower than training a single unified denoising model. Additionally, we introduce Proxy-based Pruning to further customize the denoising models. This method transforms the pruning problem of diffusion models into a multi-round decision-making problem, enabling precise pruning of diffusion models. Our experiments validate the effectiveness of TDC Training, demonstrating improvements in FID of 1.5 on ImageNet64 compared to original IDDPM, while saving about 20\% of computational resources. 
\end{abstract}


\section{Introduction}
\label{sec:intro}

Generative models, such as generative adversarial networks (GANs) \cite{b1,b2,b3,b4}, flows \cite{b5}, autoregressive models \cite{b6}, and variational autoencoders (VAEs) \cite{b7}, have showcased profound capabilities across diverse applications. Among these, diffusion models \cite{b8} represent the forefront of generative modeling, achieving notable success in areas like image generation \cite{b9,b10,b11,b12,b13,b14,b15}, super-resolution \cite{b16}, and video synthesis \cite{b17,b18}. Diffusion models consist of thousands of timesteps, employing a unified denoising model to predict noise across different timesteps and noise levels. This denoising model undertakes a series of tasks to gradually denoise, starting from the random noise at timestep $T$, and eventually arriving at the target data at timestep $0$.

\begin{figure}[t!]
  \centering
\includegraphics[width=0.47\textwidth]{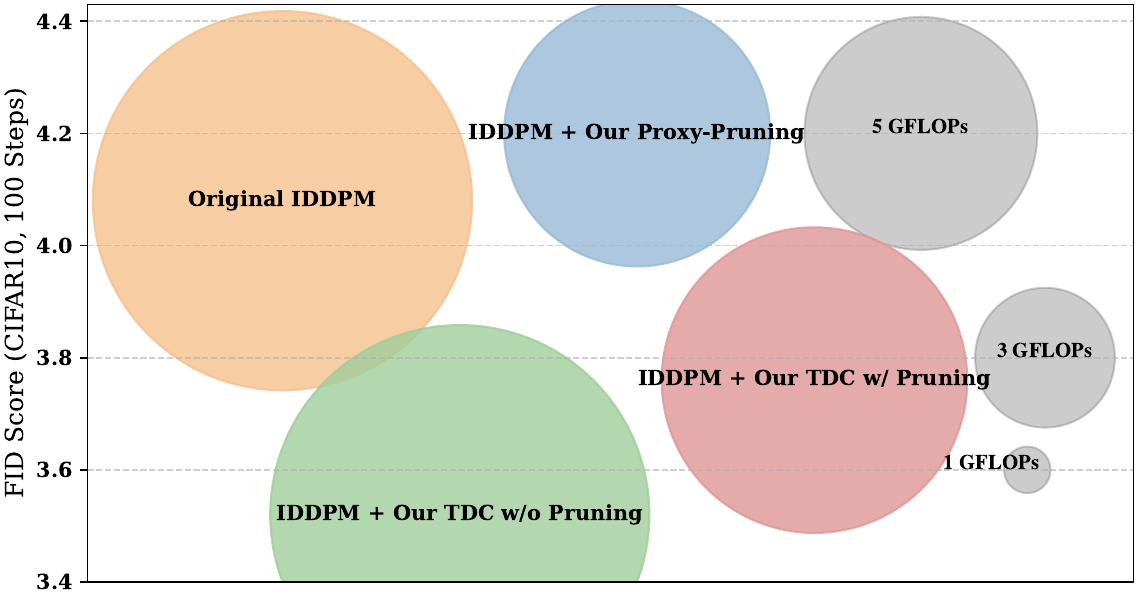}
  \caption{Visualization of Diffusion Model Performance: Circle sizes represent computational costs (GFLOPs) while vertical positioning indicates FID scores.}
  \label{first_fig}
\end{figure}

However, there are significant differences at each timestep, which lead to substantial variations in denoising tasks across different timesteps. The attempt to have a single denoising model cover all timesteps leads to slower convergence, increased training overhead, and compromised performance \cite{b53}. From timestep $T$ to timestep $0$, the data distribution gradually transforms from a Gaussian noise distribution to the target data distribution. Due to the significant differences in data distributions between different timesteps, requiring a single model to adapt to so many distributions imposes a significant capacity burden and training conflicts.
Aside from differences in data distributions, there are also significant variations in task difficulty. Consider two noisy latents, one with minor noise, revealing clear original image content, and another with substantial noise, containing limited original information. It is evident that the denoising model can more easily adapt to the former. Using models of the same capacity for each timestep can lead to redundancy in model capability for less challenging timesteps or insufficient capability for more challenging timesteps. 

Therefore, to maximize the performance of diffusion models, a more reasonable approach is to allocate different models with appropriate capabilities and capacity based on the characteristics and difficulty of the task at each timestep. To achieve this, methods like \cite{b60,b61,b62} use completely distinct denoising models for each timestep. However, this solution introduces a significant drawback – training too many models incurs prohibitively high computational costs and potentially diminishes synergy between timesteps. \cite{b63} proposes to gradually transfer models trained on easier timesteps to harder timesteps. But the performance of this method is poor due to the lack of sufficient training on target timesteps.

\begin{figure*}[t!]
  \centering
  \includegraphics[width=\textwidth]{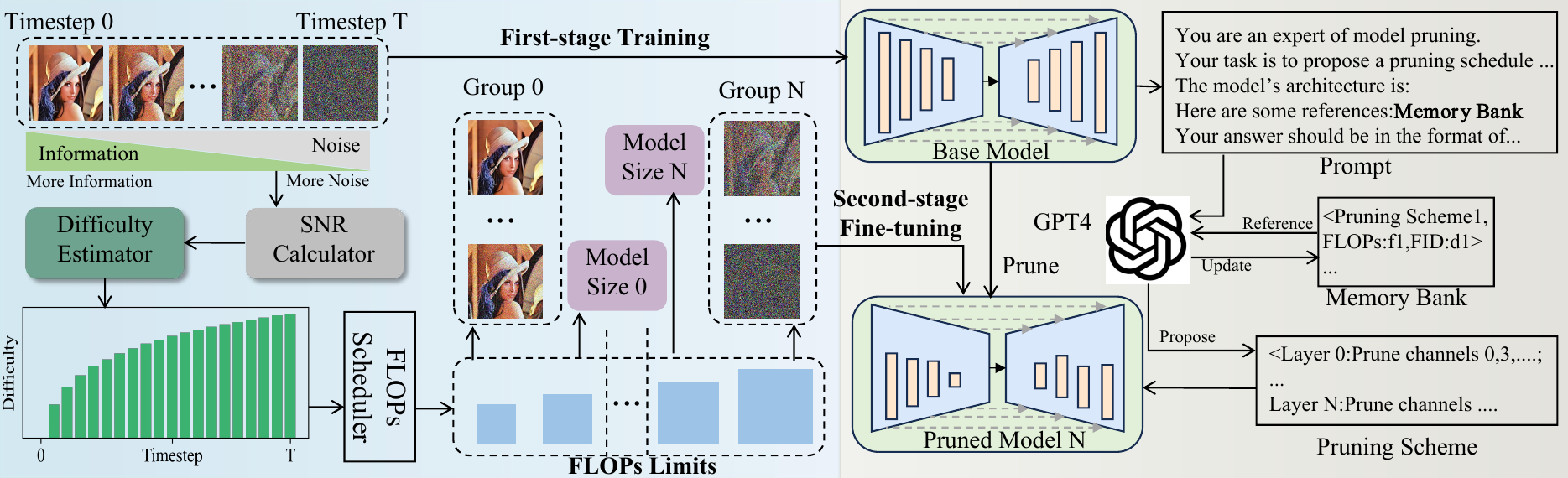}
  \caption{\textbf{Pipeline of Our TDC Training Strategy.} First, SNR for each timestep is calculated to estimate the difficulty of the denoising task. Timesteps are then grouped based on task difficulty, and model capacity is allocated accordingly. During training, a base model covering all timesteps is trained in the first phase. In the second phase, for each group, Proxy-based Pruning is applied to the base model according to the allocated model capacity, and then fine-tuning is performed on the timesteps within each group to obtain specialized models for each group.}
  \vspace{-14pt}
  \label{main_fig}
\end{figure*}

In this paper, we propose \textbf{T}wo-stage \textbf{D}ivide-and-\textbf{C}onquer training strategy termed TDC Training, which can allocate specialized models to each timestep while reducing training costs. Recognizing that adjacent timesteps share similar data distributions and noise levels, rather than training individual models for each timestep, in TDC Training, we propose to group timesteps with similar tasks together, each group sharing a denoising model. The number of models is reduced from thousands to just a few to several dozen. To further manage the training cost associated with multiple denoising models, we split the training process into two stages. Initially, a base model is trained across all timesteps. Subsequently, this model is distributed and fine-tuned independently on each timestep group. In this way, TDC Training effectively alleviates conflicts between distinct timesteps, while sharing many overlapping computational costs. 

We also propose Proxy-based Pruning for further model custimization. To ensure that each group's denoising model has the appropriate model capacity, we performed varying degrees of Proxy-based Pruning on the base model according to task difficulty before the second stage of training. Proxy-based Pruning leverages the capabilities of GPT-4 \cite{b35}, a cutting-edge general-purpose language model known for its powerful abilities in data analysis and complex algorithm design. We use it as a proxy, treating the importance evaluation in pruning as an iterative decision-making task. By inputting structured prompts that include model architecture, pruning requirements, and other information, proxy can select redundant parameters for pruning. During the iterative decision-making process, continuous feedback on pruned models' performance allows us to build a memory bank of past pruning schemes, thereby further enhancing the accuracy of the decision-making process. 

\section{Method}

\subsection{Unequal Timesteps in Denoising Capacity}
Owing to the unique design of our method, where we segregate training across different timesteps and allocate denoising models of varying capacities to different timesteps, it becomes imperative for us to first investigate the fundamental differences in the denoising tasks across these timesteps. 

The distribution of 
$\mathbf{x}_t$ can be expressed as:
\begin{equation}
\mathbf{x}_t=\sqrt{\overline{\alpha}_t}\mathbf{x}_0+\sqrt{1-\overline{\alpha}_t}\mathbf{\epsilon}.
\label{eq6}
\end{equation}

With the increase in timesteps, $\overline{\alpha}_t$ gradually increases. The difference in distribution between timesteps is the main source of difference in the denoising tasks. Adjacent timesteps have similar input distributions, rendering their tasks quite comparable. Denoising task changes progressively along timesteps.

Moreover, the distribution of $\mathbf{x}_t$ can be regarded as a linear combination of the primary signal $\mathbf{x}_0$ and random noise $\mathbf{\epsilon}$. Considering the preceding parameter as the amplitude of the signal, we can compute its Signal-to-Noise Ratio (SNR).
 \begin{equation}
SNR = 10 \log_{10}\left(\frac{\overline{\alpha}_t}{1-\overline{\alpha}_t}\right).
\label{eq7}
\end{equation}

\vspace{-5pt}

It's evident that the task difficulty differences across timesteps mainly arise from variations in SNR. Timesteps with low SNR contain less information, making the denoising task challenging. And a larger model should be employed. Thus, we can leverage the SNR to evaluate task difficulty, subsequently allocating models of varying sizes to different timesteps. Assuming the minimum FLOPs of models is $k\mathcal{F}$, and the maximum is $\mathcal{F}$, the FLOPs of models for different time steps can be determined by normalizing negative SNR values and then mapping them to the range $[k\mathcal{F},\mathcal{F}]$.
\begin{equation}
\begin{aligned}
&S_{n} = \frac{-SNR-mean(-SNR)}{std(-SNR)}\\
&FLOPs =k\mathcal{F}+\frac{(1-k)(S_{n}-min(S_{n}))}{max(S_{n})-min(S_{n})} \mathcal{F}
\label{eq8}
\end{aligned}
\end{equation}

\begin{figure*}[t!]
  \centering
  \includegraphics[width=\textwidth]{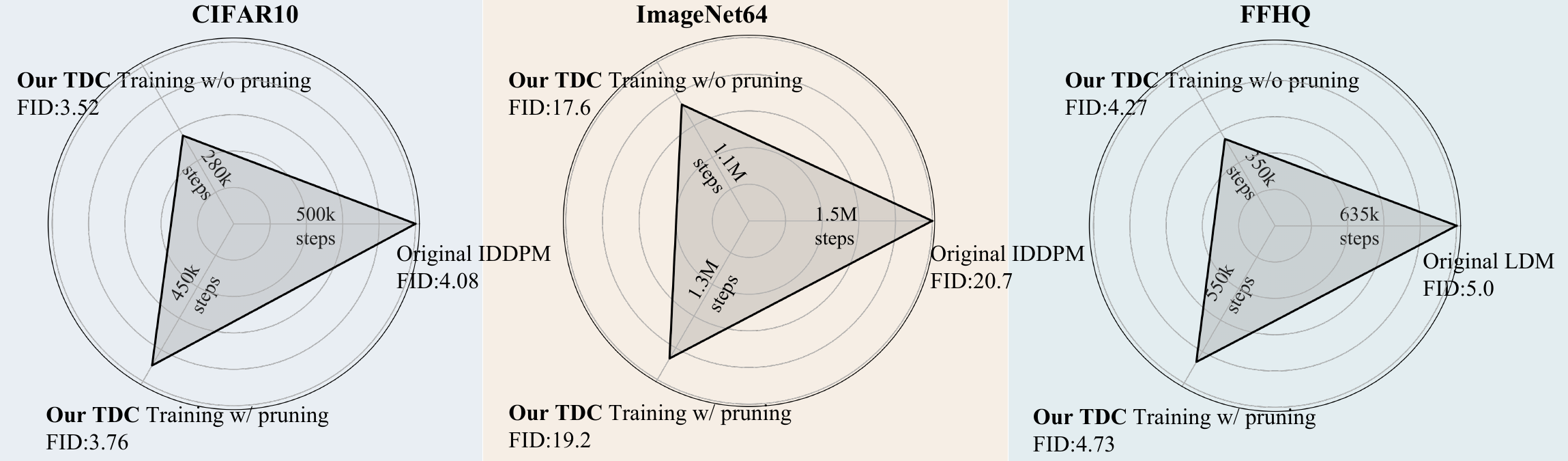}
  \caption{Comparison of FID and Training Steps Across Different Training Strategies}
  \vspace{-15pt}
  \label{img_noise}
\end{figure*}

\subsection{Progressive FLOPs Allocation with Grouped Steps}
From the previous analysis, each timestep requires a different level of model capacity and scale. But the cost of dedicating a separate model for each timestep and training it from scratch is clearly prohibitive. As adjacent timesteps share similar characteristics and levels of difficulty, a logical approach is to partition all timesteps into $\mathcal{N}$ groups based on the difficulty of the task, with each group sharing a denoising model. Given that the range of FLOPs across all timesteps is $[k\mathcal{F},\mathcal{F}]$, setting equal intervals for FLOPs between each group, the FLOPs upper limits for each group would be:
\begin{equation}
FLOPs_g(i)=(\frac{i}{\mathcal{N}}+\frac{\mathcal{N}-i}{\mathcal{N}}\times k)\mathcal{F},\quad 0\le i<N.
\label{eq14}
\end{equation}
After determining the upper FLOP limits for each group, we can allocate timesteps into N distinct groups.
\begin{equation}
\begin{aligned}
&\mathcal{T}(i)=\{t|v(i)< FLOPs(t)\le w(i)\},\quad0\le i<N\\
&v(0)=0,v(i)=FLOPs_g(i-1),\quad0<i<N\\
&w(i)=FLOPs_g(i),\quad0\le i<N,\\
\end{aligned}
\label{eq15}
\end{equation}

where $v(i)$ and $w(i)$ represent the lower and upper FLOPs limits for the $i^{th}$ group, and $\mathcal{T}(i)$ denotes the set of timesteps included in the $i^{th}$ group. In this way, we also name our obtained models as \textit{progressive diffusion models} since they are allocated with progressive FLOPs over timesteps.

\subsection{TDC Training for Progressive Diffusion Models}

While we have partitioned the timesteps into $\mathcal{N}$ groups, reducing the number of models from $\mathcal{T}$ to $\mathcal{N}$, it remains impractical to design separate denoising models meeting FLOPs constraints for each group and train them from scratch. Therefore, we propose our two-stage divide-and-conquer Training Strategy for diffusion models. 

\textbf{Two-stage Training.} In the first stage, we start by training a base denoising model across all timesteps.
\begin{equation}
\mathcal{U}_{base}^*=\mathop{\arg \min}_{\theta}~\mathop{Loss}(\mathcal{U}_{base}(\mathcal{T}, \mathcal{D}, \mathcal{S}))
\label{eq16}
\end{equation}

\begin{algorithm}[t!]
\caption{TDC Training}
\begin{algorithmic}[1]
\State \textbf{Input:} Total timesteps $\mathcal{T}$, dataset $\mathcal{D}$, settings $\mathcal{S}$, memory bank $\mathcal{M}$, step groups $\mathcal{N}$, iterative rounds $\mathcal{R}$

\State \textbf{Preparation: Group Timesteps by Difficulty}
\State Determine FLOPs upper limits $\mathcal{F}$ for each group
\State Allocate timesteps to $\mathcal{N}$ groups based on limits $\mathcal{F}$ 

\State \textbf{Stage 1: Train Base Model}
\State Update $\mathcal{U}_{base}$ to minimize loss over $\mathcal{D}$ for $t \in \mathcal{T}$

\State \textbf{Stage 2: Group-Based Pruning and Fine-Tuning}
\For{each group $i = 0$ to $\mathcal{N}-1$}
    \While{Iterative rounds not met $\mathcal{R}$}
        \State Prune $\mathcal{U}_{base}$ to conform to group $i$'s limit $\mathcal{F}$
        \State Update $\mathcal{M}$ with pruning scheme and performance
    \EndWhile
    \State Fine-tune $\mathcal{U}^*(i)$ for group $i$
\EndFor
\end{algorithmic}
\end{algorithm}

After acquiring the base denoising model, denoted as $\mathcal{U}_{base}$, in the second stage, we commence by pruning the base model. This process is aimed at deriving the optimal sub-model for each group that conforms to the predefined FLOPs constraint. If we consider the importance evaluation mechanisms as a proxy, this process can be described as follows:
\begin{equation}
\begin{aligned}
&\mathcal{U}(i)= \mathcal{U}_{base}^*-\theta'(i),0\le i<N\\
&\theta'(i)=Proxy(\theta,\mathcal{T}(i),\mathcal{D},\mathcal{S},\mathcal{M},FLOPs_g(i)),\\
\end{aligned}
\label{eq17}
\end{equation}
where $\theta'(i)$ denotes the structured parameters to be pruned within the $i^{th}$ group, while $\mathcal{U}(i)$ represents the optimal subnetwork obtained for the $i^{th}$ group.

\begin{table*}[t!]
\centering
\renewcommand{\arraystretch}{1}
\small
\caption{Overall Results on IDDPM and latent diffusion}
\label{table1}
\setlength{\tabcolsep}{14pt} 
\begin{tabular}{c|cccccc} 
\toprule
\textbf{Dataset} & \textbf{Model} & \textbf{Training Strategy} & \textbf{FID} & \textbf{FLOPs} & \textbf{Params} & \textbf{Train Steps} \\
\midrule
\multirow{7}{*}{\shortstack{CIFAR10\\(100 DDIM)}} & WaveDiff\cite{b47} & Original Strategy & 4.87 & 6.06G & 38.2M & 800k \\
& PNDM\cite{b50} & Original Strategy & 4.10 & 6.06G & 38.2M & 800k\\
& DDPM\cite{b8} & Original Strategy & 4.19 & 6.06G & 38.2M & 800k\\
& IDDPM\cite{b34} & Original Strategy & 4.08 & 8.14G & 52.5M & 500k\\
& \cellcolor{gray!20}IDDPM\cite{b34} & \cellcolor{gray!20}\textbf{TDC Training w/o pruning} & \cellcolor{gray!20}3.52 & \cellcolor{gray!20}8.14G & \cellcolor{gray!20}52.5M & \cellcolor{gray!20}280k\\
& \cellcolor{gray!20}IDDPM\cite{b34} & \cellcolor{gray!20}\textbf{TDC Training} & \cellcolor{gray!20}3.76 & \cellcolor{gray!20}6.56G & \cellcolor{gray!20}48.7M& \cellcolor{gray!20}450k\\
\cline{1-7}
\multirow{4}{*}{\shortstack{ImageNet64\\(100 DDIM)}} & IDDPM\cite{b34} & Original Training & 20.7 & 39.3G & 121M & 1.5M\\ 
& \cellcolor{gray!20}IDDPM\cite{b34} & \cellcolor{gray!20}\textbf{TDC Training w/o pruning} & \cellcolor{gray!20}17.6 & \cellcolor{gray!20}39.3G & \cellcolor{gray!20}121M & \cellcolor{gray!20}1.1M\\ 
& \cellcolor{gray!20}IDDPM\cite{b34} & \cellcolor{gray!20}\textbf{TDC Training} & \cellcolor{gray!20}19.2 & \cellcolor{gray!20}32.7G & \cellcolor{gray!20}103M & \cellcolor{gray!20}1.3M\\ 
\cline{1-7}
\multirow{6}{*}{\shortstack{FFHQ\\(200 DDIM)}} & DDPM\cite{b8} & Original Training & 8.4 & 248.7G & 113.7M & 800k\\
& P2\cite{b58} & Original Training & 7.0 & 248.7G & 113.7M & 800k\\
& LDM\cite{b59} & Original Training & 5.0 & 96.1G & 274.1M & 635k\\
& \cellcolor{gray!20}LDM\cite{b59} & \cellcolor{gray!20}\textbf{TDC Training w/o pruning} & \cellcolor{gray!20}4.27 & \cellcolor{gray!20}96.1G & \cellcolor{gray!20}274.1M & \cellcolor{gray!20}350k\\
& \cellcolor{gray!20}LDM\cite{b59} & \cellcolor{gray!20}\textbf{TDC Training} & \cellcolor{gray!20}4.73 & \cellcolor{gray!20}77.2G & \cellcolor{gray!20}231.0M& \cellcolor{gray!20}550k\\
\bottomrule
\end{tabular}
\vspace{-20pt} 
\end{table*}

After pruning, we fine-tune each model on the timesteps within the group to achieve optimal performance.
\begin{equation}
\mathcal{U}^*(i)=\mathop{\arg \min}_{\theta}~\mathop{Loss}(\mathcal{U}(i))(\mathcal{T}(i), \mathcal{D}, \mathcal{S})).
\label{eq18}
\end{equation}

\textbf{Proxy-based Pruning.} To appropriately allocate model sizes for different groups, we prune the base model before the second stage of training. The objective of model pruning is to reduce non-essential parameters or structures while striving to maintain optimal model performance. Given a dataset $\mathcal{D}$, a base denoising model $\mathcal{U}$, target timesteps $\mathcal{T}$, other settings $\mathcal{S}$, and an upper limit of FLOPs denoted as $f$, the pruning procedure for the diffusion model is as follows:
\begin{equation}
\begin{aligned}
&{\mathcal{U}_{\theta-\theta'}}^* = \mathop{\arg \min}_{\theta' \in \theta}~\mathop{Loss}(\mathcal{U}_{\theta-\theta'}(\mathcal{T}, \mathcal{D}, \mathcal{S})) \\
&\text{s.t.}\quad \text{FLOPs}(\mathcal{U}_{\theta-\theta'}) \leq f,\\
\label{eq9}
\end{aligned}     
\end{equation}
\vspace{-20pt}

where $\theta$ denotes the parameters of the base denoising model, while $\theta'$ indicates the parameters to be pruned. The crux of model pruning lies in evaluating and ranking the importance of parameters or structures. If we represent the process of determining the least significant parameters through importance evaluation as $\mathcal{P}$, then \cref{eq9} can be updated as follows:
\begin{equation}
\begin{aligned}
&\theta'=\mathcal{P}(\theta,\mathcal{T},\mathcal{D},\mathcal{S}),\ \theta' \subseteq \theta\\
&\text{s.t.}\quad \text{FLOPs}(\mathcal{U}_{\theta-\theta'}) \leq f.\\
\label{eq10}
\end{aligned}     
\end{equation}

\vspace{-15pt}

Since traditional important evaluation methods are not suitable for scenarios where diffusion models are sensitive to small changes in output, we propose to utilize GPT-4's powerful understanding ability over architectures, and leverage it as a proxy for the assessment of model parameter importance, aiding in the identification of the least important group of parameters. Besides, we also introduce a memory bank to store the performance of each pruned model so that it can serve as a feedback and the whole pruning is implemented in an iterative manner for boosted performance, i.e.
\begin{equation}
\begin{aligned}
&\theta'=GPT(\theta,\mathcal{T},\mathcal{D},\mathcal{S},\mathcal{M}),\ \theta' \subseteq \theta\\
&\text{s.t.}\quad \text{FLOPs}(\mathcal{U}_{\theta-\theta'}) \leq f,\\
\label{eq12}
\end{aligned}     
\end{equation}
where $\mathcal{M}$ denotes the memory bank, which saves the pruning scheme from each pruning iteration along with the performance metrics of the model post-pruning.  Its update process is as follows:
\begin{equation}
\mathcal{M}_{i+1}=\mathcal{M}_i \cup \left(\theta'_i, Loss(\mathcal{U}_{\theta-\theta'_i}(\mathcal{T},\mathcal{D},\mathcal{S}))\right).
\label{eq13}
\end{equation}

In our Proxy-based Pruning, all information, including the model architecture, is input into GPT-4 in the form of a structured prompt, and the pruning scheme is obtained by specifying the output format.

\begin{figure}[t!]
  \centering
 \centering
    \includegraphics[width=0.47\textwidth]{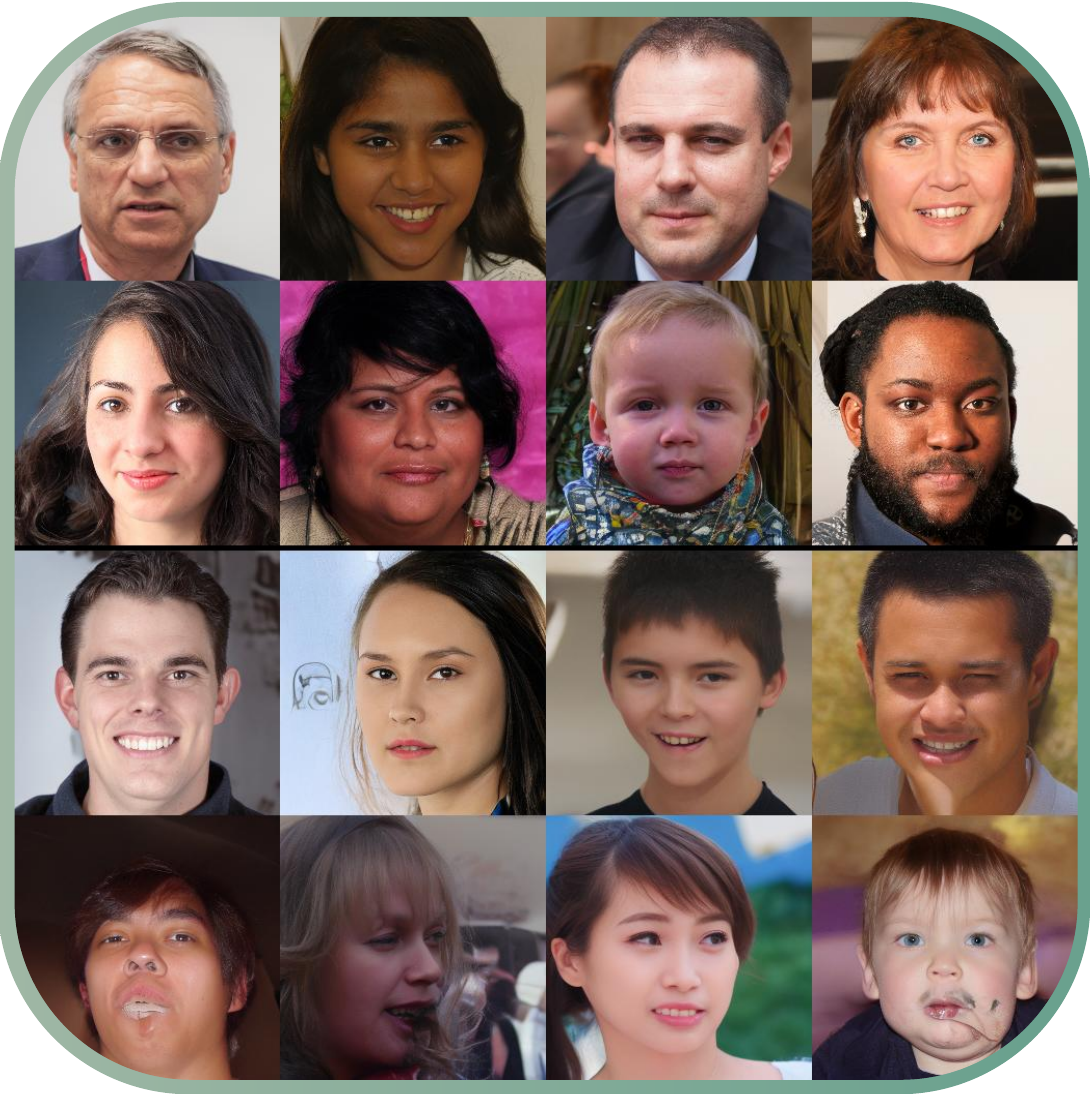}
  \caption{Sample images of LDM on FFHQ with (top) and without (bottom) our TDC Training(100 sampling steps).}
  \label{fig_sample}
\end{figure}

\section{Experiments}

In our experiments, we employed representative diffusion models IDDPM and latent diffusion as baselines, conducting experiments on three datasets of varying resolutions: CIFAR10, ImageNet64, and FFHQ. We employed 100-step fast sampling in all experiments of IDDPM and 200-step fast sampling in all experiments of latent diffusion. In all experiments, the number of groups is set to 10, and the minimum FLOPs constraint $k$ is set to 0.5.

\begin{table}[t!]
    \centering
    \renewcommand{\arraystretch}{1}
    \setlength{\tabcolsep}{18pt}
    \caption{FID Scores with Different FLOPs Schedule on CIFAR10}
    \label{table3}
    \begin{tabularx}{\columnwidth}{c|c|c|c}
        \toprule
        Schedule & FID & FLOPs & Params\\
        \midrule
        \textbf{Our} & 3.76 & 6.56G & 48.7M\\
        Constant & 3.91 & 6.79G & 49.5M \\
        Uni-increasing & 3.88 & 6.85G & 43.2M \\
        Uni-decreasing & 4.11 & 6.21G & 45.1M \\
        \bottomrule
    \end{tabularx}
    \vspace{-5pt}
\end{table}

\begin{table}[t!]
    \centering
    \setlength{\tabcolsep}{10.5pt}
    \caption{FID Changes for Different Group Numbers on CIFAR10}
    \label{tab:fid_changes}
    \begin{tabularx}{\columnwidth}{c|ccccc}
        \toprule
        Group  Number & 4 & 8 & 10 & 15 & 20 \\
        \midrule
        \makecell{FID} & 3.62 & 3.56 & 3.52 & 3.56 & 3.60 \\
        \bottomrule
    \end{tabularx}
    \vspace{-5pt}
\end{table}

\begin{table}[t!]
    \centering
    \caption{Time Consuming(ks)}
    \label{tab:model_training_time}
    \begin{tabularx}{\columnwidth}{c|c|c|c}
        \toprule
         GPT-4 Inference & Finetuning & Our Total Time & OMS-DPM \cite{b60} \\
        \midrule
         0.078 & 190 & 342 & $>$ 3800 \\
        \bottomrule
    \end{tabularx}
\end{table}

\subsection{Experiments on TDC Training}
\label{sec:exp1}

\begin{table*}[t!]
\centering
\small
\caption{Performance of Different Pruning Methods}
\label{table4}
\renewcommand{\arraystretch}{1} 
\setlength{\tabcolsep}{10.5pt} 
\begin{tabular}{c|cccc|cccc} 
\toprule
\textbf{Pruning Method} & \multicolumn{4}{c|}{\textbf{IDDPM (100 DDIM) CIFAR10}} & \multicolumn{4}{c}{\textbf{LDM (200 DDIM) FFHQ}}\\
\midrule
 & \textbf{Params} & \textbf{FLOPs} & \textbf{FID}  & \textbf{Train Steps} & \textbf{Params} & \textbf{FLOPs} & \textbf{FID}  & \textbf{Train Steps}\\
\hline
Pretrained & 52.5M & 8.14G & 4.08 & 500K & 274.1M & 96.1G & 5.0 & 635k\\
Scratch Training & 38.7M & 5.71G & 89.1 & 25K & 182.7M & 66.3G & 85.0 & 30k\\
Scratch Training & 38.7M & 5.71G & 16.8 & 100K & 182.7M & 66.3G & 12.7 & 200k\\
Scratch Training & 38.7M & 5.71G & 5.1 & 500K & 182.7M & 66.3G & 6.9 & 635k\\
Random Pruning & 40M & 6.02G  & 4.9 & 25K & 190M & 63.2G & 6.5 & 30k\\
Magnitude Pruning\cite{b56} & 40M &  6.29G & 4.6 & 25K & 190M & 70.8G & 6.2 & 30k\\
Taylor Pruning\cite{b57} & 40M &  5.90G & 4.7 & 25K & 190M & 68.7G & 6.2 & 30k\\
Diff Pruning\cite{b52} & 40M & 5.88G & 4.4 & 25K & 190M & 66.8G & 6.1 & 30k\\
\rowcolor{gray!20}\textbf{Proxy Pruning} & 38.7M & 5.71G & \textbf{4.2} & 25K & 182.7M & 66.3G & \textbf{5.8} & 30k\\
\bottomrule
\end{tabular}
\vspace{-3mm} 
\end{table*}

\begin{table*}[t!]
\centering
\caption{Performance with Different Pruning Methods}
\label{tab:model_comparison}
\renewcommand{\arraystretch}{1.05} 
\setlength{\tabcolsep}{18.5pt} 
\begin{tabular}{c|ccc|ccc} 
\toprule
\textbf{Training Strategy} & \multicolumn{3}{c|}{\textbf{IDDPM (100 DDIM) CIFAR10}} & \multicolumn{3}{c}{\textbf{LDM (200 DDIM) FFHQ}} \\
\midrule
 & \textbf{FID} & \textbf{FLOPs} & \textbf{Parameters} & \textbf{FID} & \textbf{FLOPs} & \textbf{Parameters} \\
\hline
Original Training & 4.08 & 8.14G & 52.5M & 5.0 & 96.1G & 274.1M \\
+TDC Training (Random) & 3.95 & 6.71G & 46.6M & 4.97 & 77.7G & 217.4M \\
+TDC Training (Magnitude) & 3.88 & 6.75G & 47.3M & 4.86 & 75.2G & 230.9M \\
+TDC Training (Taylor) & 3.90 & 6.20G & 45.3M & 4.89 & 77.8G & 215.4M \\
+TDC Training (Diff-Pruning) & 3.85 & 6.82G & 46.1M & 4.81 & 75.7G & 223.3M \\
\rowcolor{gray!20}+TDC Training (\textbf{Proxy}) & \textbf{3.76} & 6.56G & 48.7M & \textbf{4.73} & 77.2G & 231.0M \\
\bottomrule
\end{tabular}
\vspace{-3mm} 
\end{table*}

Firstly, we conducted comparative experiments to verify the efficacy of TDC Training with or without pruning. In TDC Training without pruning, we allow each specialized model to inherit the architecture of the base model without pruning. The results are shown in \cref{table1}. Taking LDM on FFHQ as an example, with the same FLOPs, TDC Training can reduce the model's FID from 5.0 to 4.27. After pruning, FLOPs are reduced from 96.1G to 77.2G, and the FID is 4.73, which is 0.27 better than the original LDM.

\subsection{Comparative Experiments of Pruning Methods}
To evaluate the effectiveness of proxy-pruning, we first examined the capability of itself. A natural baseline was to train a smaller network from scratch. We also compared our method with several common general-purpose pruning methods as well as Diff-Pruning, the pruning method specifically designed for diffusion models. The results are shown in \cref{table4}. Furthermore, we explored the impact of replacing pruning algorithms on the overall effect of the TDC Training. The experimental results are shown in \cref{tab:model_comparison}. Compared to the original IDDPM, TDC Training with Proxy-based Pruning reduces the FLOPs from 8.14G to 6.56G, while lowering the FID from 4.08 to 3.76.

\vspace{-5pt}

\subsection{Comparative Analysis: Single-Stage vs. Two-Stage Training Strategies}

\begin{figure}[t!]
\vspace{-8pt}
  \centering
 \centering
    \includegraphics[width=0.45\textwidth]{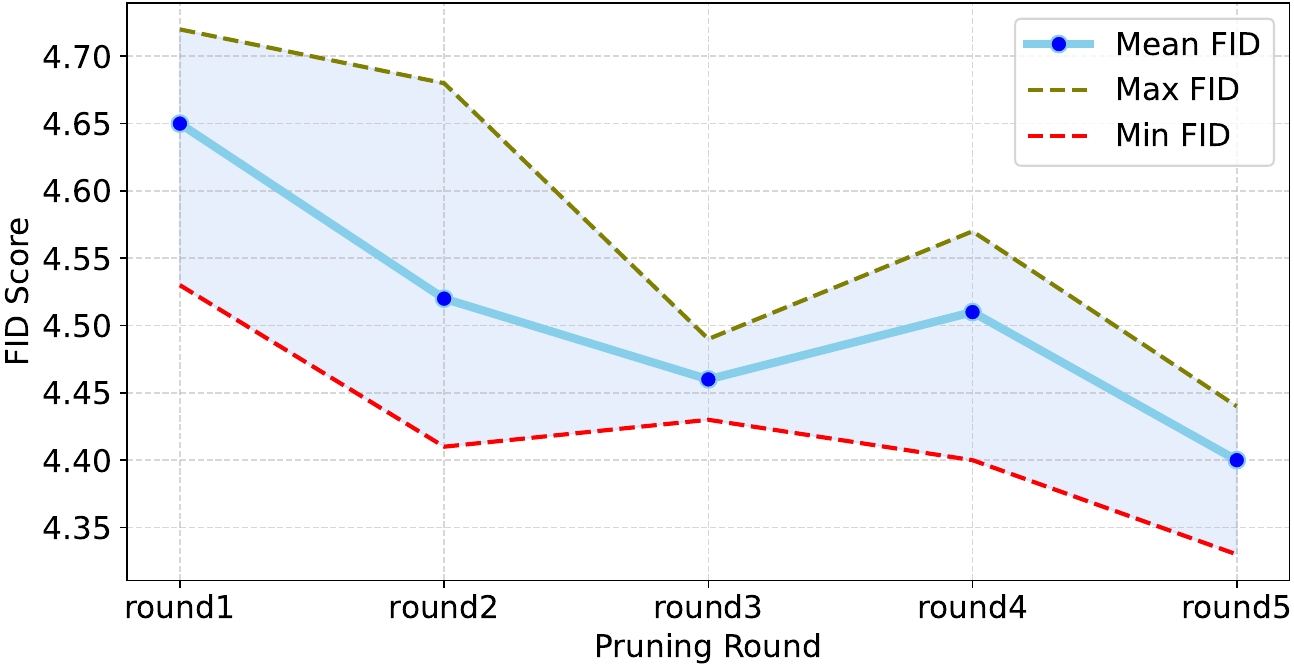}
  \caption{Mean-std Curve over Pruning Rounds.}
  \label{fig5}
\end{figure}

\begin{figure}[t!]
\vspace{-3pt}
  \centering
 \centering
    \includegraphics[width=0.45\textwidth]{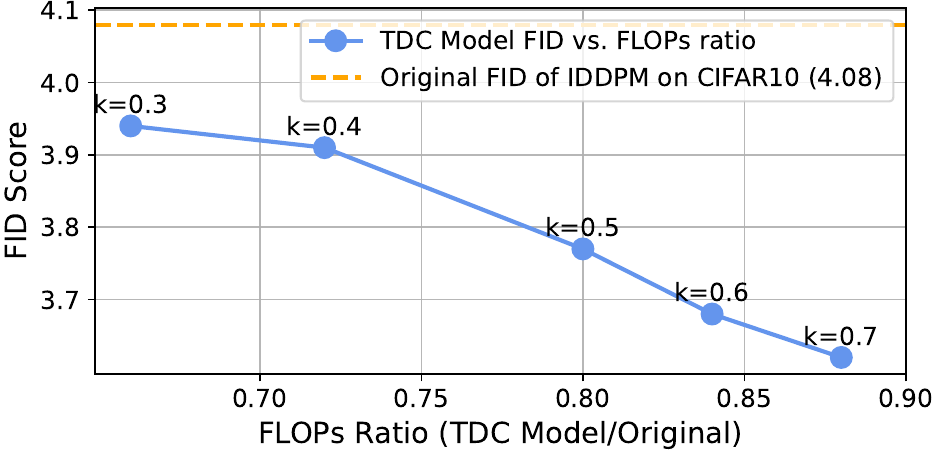}
  \caption{Ablation Study of $k$ on IDDPM.}
  \vspace{-2.5pt}
  \label{img_k_iddpm}
\end{figure}

\begin{figure}[t!]
\vspace{-8pt}
  \centering
 \centering
    \includegraphics[width=0.4\textwidth]{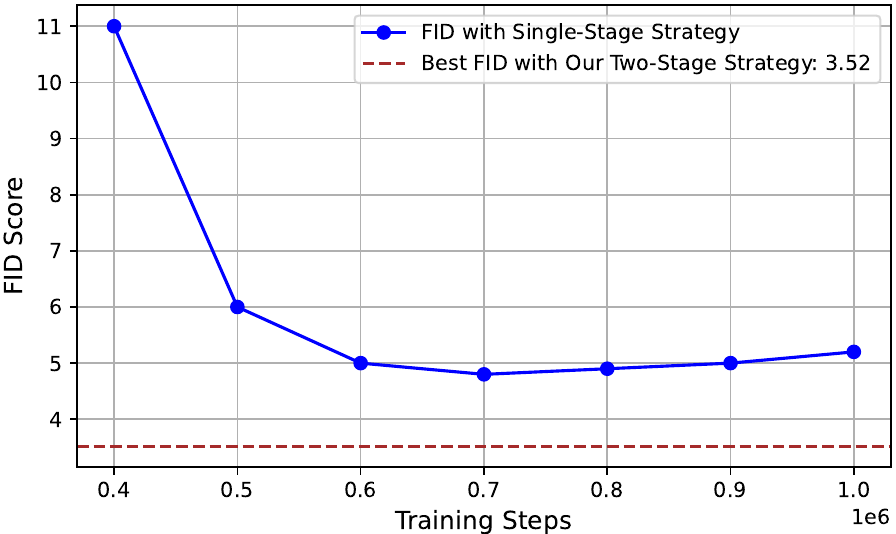}
  \caption{Comparative FID Scores: Single-Stage vs. Two-Stage Strategies in IDDPM on CIFAR10.}
  \label{fig_append1}
\end{figure}

To further validate the effectiveness of the two-stage training in our TDC training strategy, we conducted an comparative experiment with a single-stage divide-and-conquer training strategy. Specifically, we maintained the same training settings on CIFAR10 as in the experiment described in \cref{sec:exp1}, but we abandoned the two-stage training. Instead, we trained models for each group from scratch. The experimental results are shown in \cref{fig_append1}.

\subsection{Stability of Proxy Pruning}

\begin{figure}[t!]
  \centering
 \centering
    \includegraphics[width=0.45\textwidth]{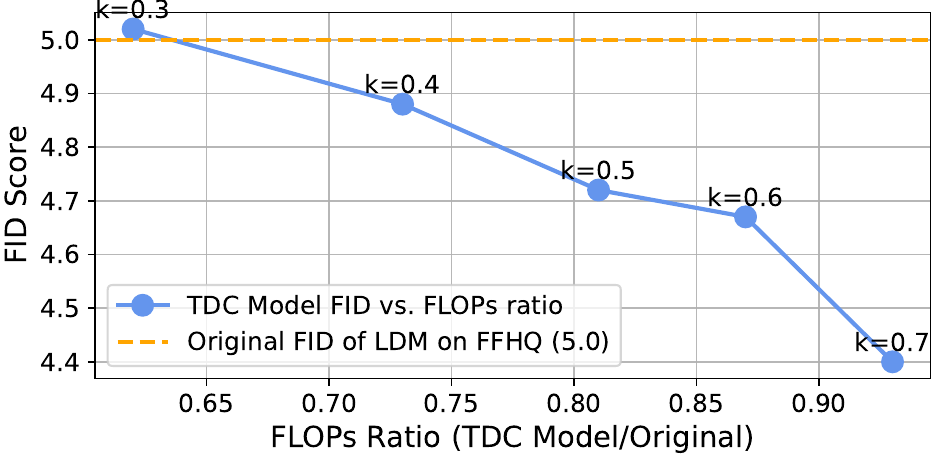}
  \caption{Ablation Study of $k$ on LDM.}
  \label{img_k_ldm}
\end{figure}

Due to the inherent variance in GPT-4's inference, which is utilized in our proxy pruning algorithm, the importance assessments obtained are not entirely consistent. Hence, designing experiments to evaluate the stability of the pruning algorithm is of great significance. In these experiments, we pruned the IDDPM on CIFAR10 with a pruning rate set at 0.7, obtaining three pruning outcomes each time. After evaluating the performance of these three sub-models, we stored the results in a memory bank for the next round of three prunings. This process was repeated for five rounds, with the results illustrated in \cref{fig5}.

\subsection{Ablation Study of FLOPs Constraint $k$}
In all previous experiments, we set the FLOPs Constraint $k$ to 0.5. This value affects both the overall computational requirements of the model and its performance. Therefore, we designed ablation studies on this parameter. Specifically, we set the value to 0.3, 0.4, 0.5, 0.6, and 0.7, and conducted experiments on both IDDPM and latent diffusion models using CIFAR10 and FFHQ datasets, respectively. The results, as shown in \cref{img_k_iddpm} and \cref{img_k_ldm}, indicate that our algorithm performs stably across different values of $k$.

\section{Conclusion}
In this paper, we introduced TDC training strategy for diffusion models. It groups timesteps based on task difficulty and characteristics, assigning customized models to different groups, thereby improving the performance of diffusion models. Additionally, it reduces training costs with two-stage training. To further customize the denoising models, we proposed Proxy-based pruning. The combination of TDC Training and the Proxy-based pruning resulted in an improvement of 0.32, 1.5 and 0.27 in FID on CIFAR10, ImageNet64 and FFHQ datasets, respectively, with an approximate 20\% reduction in the model's computational requirements. 

\bibliographystyle{IEEEbib}
\bibliography{icme2025references}

\end{document}